\documentclass[sigconf]{acmart}

\usepackage{amssymb}
\usepackage{threeparttable}
\usepackage{algorithm}
\usepackage{algorithmic}
\usepackage{longtable}
\usepackage{multirow}

\AtBeginDocument{%
  \providecommand\BibTeX{{%
    \normalfont B\kern-0.5em{\scshape i\kern-0.25em b}\kern-0.8em\TeX}}}

\AtBeginDocument{%
    \providecommand\BibTeX{{%
    Bib\TeX}}}

\copyrightyear{2024}
\acmYear{2024}
\setcopyright{rightsretained}
\acmConference[MCHM '24]{Proceedings of the 1st International Workshop on Multimedia Computing for Health and Medicine}{October 28-November 1, 2024}{Melbourne, VIC, Australia}
\acmBooktitle{Proceedings of the 1st International Workshop on Multimedia Computing for Health and Medicine (MCHM '24), October 28-November 1, 2024, Melbourne, VIC, Australia}
\acmDOI{10.1145/3688868.3689195}
\acmISBN{979-8-4007-1195-4/24/10}

\makeatletter
\gdef\@copyrightpermission{
 \begin{minipage}{0.3\columnwidth}
  \href{https://creativecommons.org/licenses/by/4.0/}{\includegraphics[width=0.90\textwidth]{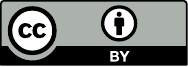}}
 \end{minipage}\hfill
 \begin{minipage}{0.7\columnwidth}
  \href{https://creativecommons.org/licenses/by/4.0/}{This work is licensed under a Creative Commons Attribution International 4.0 License.}
 \end{minipage}
 \vspace{5pt}
}
\makeatother

\acmSubmissionID{6}

\begin{document}

\title{A Multimodal Object-level Contrast Learning Method for Cancer Survival Risk Prediction}

% \affiliation{%
\author{Zekang Yang}
\affiliation{%
  \institution{ Beijing Key Laboratory of Mobile Computing and Pervasive Device, Institute of Computing Technology,
 Chinese Academy of Sciences \&
 University of Chinese Academy of
 Sciences}   \city{Beijing}
   \country{China}}
  % \streetaddress{1 Th{\o}rv{\"a}ld Circle}}
\email{yangzekang22s@ict.ac.cn}

\author{Hong Liu}
\authornote{Corresponding author.}
\affiliation{%
  \institution{ Beijing Key Laboratory of Mobile Computing and Pervasive Device, Institute of Computing Technology,
 Chinese Academy of Sciences}\city{Beijing}
   \country{China}}
  % \streetaddress{1 Th{\o}rv{\"a}ld Circle}}
\email{hliu@ict.ac.cn}

\author{Xiangdong Wang}
\affiliation{%
  \institution{ Beijing Key Laboratory of Mobile Computing and Pervasive Device, Institute of Computing Technology,
 Chinese Academy of Sciences}\city{Beijing}
   \country{China}}
  % \streetaddress{1 Th{\o}rv{\"a}ld Circle}}
\email{xdwang@ict.ac.cn}

% \author{Anonymous Authors}
%% You do not have to enter your paper ID


\begin{abstract}
Computer-aided cancer survival risk prediction plays an important role in the timely treatment of patients. This is a challenging weakly supervised ordinal regression task associated with multiple clinical factors involved such as pathological images, genomic data and etc. In this paper, we propose a new training method, multimodal object-level contrast learning, for cancer survival risk prediction. First, we construct contrast learning pairs based on the survival risk relationship among the samples in the training sample set. Then we introduce the object-level contrast learning method to train the survival risk predictor. We further extend it to the multimodal scenario by applying cross-modal constrast. Considering the heterogeneity of pathological images and genomics data, we construct a multimodal survival risk predictor employing attention-based and self-normalizing based nerural network respectively. Finally, the survival risk predictor trained by our proposed method outperforms state-of-the-art methods on two public multimodal cancer datasets for survival risk prediction. The code is available at \hyperref[https://github.com/yang-ze-kang/MOC]{https://github.com/yang-ze-kang/MOC}.
\end{abstract}


\begin{CCSXML}
<ccs2012>
<concept>
<concept_id>10002950.10003648.10003688.10003694</concept_id>
<concept_desc>Mathematics of computing~Survival analysis</concept_desc>
<concept_significance>500</concept_significance>
</concept>
<concept>
<concept_id>10010405.10010444</concept_id>
<concept_desc>Applied computing~Life and medical sciences</concept_desc>
<concept_significance>500</concept_significance>
</concept>
</ccs2012>
\end{CCSXML}

\ccsdesc[500]{Applied computing~Life and medical sciences}
\ccsdesc[500]{Mathematics of computing~Survival analysis}

\keywords{Pathological images, Multimodal assisted diagnosis, Survival prediction, Contrast Learning}

\maketitle

\section{Introduction}
Cancer survival risk prediction refers to using computer technology to mine biomarkers from the patients' clinical data, such as gene expression, gene mutation or pathological images to estimate patients' survival risk. The more accurate of the estimation, the more timely and personalized treatment patients can get.

In survival analysis, one important issue that needs to be considered is censoring problem, which means subjects are censored if they are not followed up or the study ends before they die or have an outcome of interest. It's challenging for effectively tackling censoring problem, which also can be treated as a challenging weakly supervised and ordinal regression task.  

Deep learning has achieved great success in many fields in recent years, and there has been emerging a lot of research using deep learning methods for survival prediction. Methodologically, the core idea of most deep survival networks is optimizing the loss function derived from the Cox proportional hazards model (Cox-based), alternatively dividing the time into serval intervals with fixed boundaries (Interval-based). Both of these methods foucus on the specification of an appropriate optimization criterion for deep survival network training. Cox-based methods \cite{cox1972regression, katzman2018deepsurv, cheerla2019deep, qiu2023deep} mainly optimize the negative log partial likelihood, which promotes overall concordance by penalizing all discordance prediction. Interval-based methods \cite{ren2019deep, zadeh2020bias, chen2020pathomic, chen2021multimodal, chen2022pan} consider discrete time intervals and model each interval using an independent output neuron, then can use the cross entropy loss function of the classification problem to optimize the network.

However, both of these methods has its own shortcomings.
% Cox-based methods place additional strong assumptions in that all samples have the same baseline hazard function. 
In the training phase, for each sample, Cox-based methods need to know the risk prediction of the samples whose ground-truth survival time is longer than it and give them penality. This makes it difficult to apply deep learning methods on histological images which require a unaffordable GPU memory footprint to apply the technology of batch size. Interval-based methods convert continuous into discrete time intervals and model each interval using an independent output neuron. This formulation overcomes the need for batched samples. However, the division of time intervals depends on the distribution of real survival risks which is often unknowable. And their consideration of concordance is coarse-grained, which degrade the performance of models for predicting fine-grained cocordance. Simultaneously, in recent years, mining the relationship between pathological images and genomics data and utilizing multimodal data for enhancing survival prediction shows great promise and significant potential \cite{chen2022pan,li2022hfbsurv, ding2023pathology, lv2022transsurv, ning2022mutual}.

To address these problems, we construct a multimodal survival risk predictor and propose a multimodal object-level contrast (MOC) learning method. 
First, we design the object-level contrast (OC) learning method, utilizing contrast learning pairs to train the survival risk predictor. This not only models fine-grained concordance, but also, in every learning iteration, sample only needs to know the prediction of one other sample.
Before training, we construct contrast learning pairs based on the ground-truth survival relationship between samples. In the training phase, each sample gets its optimization direction by observing its partner which allows the model to learn the concordance of the survival relationships between samples. We then extend it to multimodal scenario through cross-modal contrast learning, proposing the multimodal object-level contrast (MOC) learning. For constructing multimodal survival predictor, we employ the attention-based neural network to process pathological images and the self-normalizing neural network to process genomic data. This helpes us eliminate the heterogeneity between modalities, allowing us to build a multimodal survival risk predictor (MSRP). The experiment results show that the MSRP trained by MOC outperforms existing SOTA methods on two multimodal cancer survival prediction datasets. The main contributions of this paper are:

\begin{itemize}
    \item We propose a new training method object-level contrast (OC) learning for cancer survival risk predictor. Different from the previous works, we use a object-level contrast learning among samples to train the survival risk predictor, which makes it convenient to apply deep learning methods and also enables for capturing fine-grained concordance. 
    \item We further extend the OC to multimodal scenarios, proposing the multimodal object-level contrast (MOC) learning for multimodal cancer surivival risk predictor. Which inspires the learning of each modality's survival predictor through cross-modal contrast, thereby enhancing the performance of the multimodal survival risk prediction model.
    \item We construct a multimodal survival risk predictor using the attention based neural network and self-normalizing neural network. We then train it using MOC and achieves the best performance compared with state-of-the-art works on two public cancer datasets.
\end{itemize}

\section{Related Work}
\textbf{Surivival Analysis Methods.} Cox \cite{cox1972regression} first proposed Cox proportion hazard model to handle censoring problem, in which hazard is parameterized as an exponential linear function. Katzman et al. \cite{katzman2018deepsurv} proposed the negative log partial likelihood, allowing a deep architecture and apply modern deep learning techniques \cite{ioffe2015batch, srivastava2014dropout, kingma2017adam}
% (i.e. regularization techniques \cite{ioffe2015batch,srivastava2014dropout}, gradient descent optimization algorithms \cite{kingma2017adam, nesterov2013gradient} and learning rate scheduling \cite{senior2013empirical})
to optimize. Qiu et al. \cite{qiu2023deep} also utilized the negative log partial likelihood to optimize their proposed multimodal model, in which they integrated pathological images and genomic data to improve survival prediction. Steck et al. \cite{steck2007ranking} showed that maximizing Cox’s partial likelihood can be understood as maximizing a lower bound on the concordance index and proposed an exponential lower bound that can achieve better performance. Ren et al. \cite{ren2019deep} turned the survival time into discrete time and trained using the cross-entropy loss function. Zadeh et al. \cite{zadeh2020bias} showed that replacing the cross-entropy loss by the negative log-likelihood loss (NLL) results in much better calibrated prediction rules and also in an improved discriminatory power. Many recent works \cite{chen2020pathomic, chen2021multimodal, chen2022pan, chen2022scaling} also adopted negative log-likelihood loss as their objective function.

\textbf{Multimodal Cancer Survival Risk Prediction Methods.} Integrating genomics and histology data to develop joint image-omic prognostic models is promising \cite{chen2020pathomic,chen2021multimodal,chen2022pan,qiu2023deep}. Chen et al. \cite{chen2021multimodal} proposed a Multimodal Co-Attention Transformer (MCAT) method that learns an interpretable, dense co-attention mapping between WSIs and genomic features formulated in an embedding space and achieved state-of-the-art performance. Chen et al. \cite{chen2022pan} proposed a deep-learning-based multimodal fusion (PORPOISE) algorithm that uses both H\&E WSIs and molecular profile features (mutation status, copy-number variation, RNA sequencing expression) to measure and explain relative risk of cancer death. Qiu et al. \cite{qiu2023deep} proposed a novel biological pathway informed pathology-genomic deep model (PONET) that integrates pathological images and genomic data not only to improve survival prediction but also to identify genes and pathways that cause different survival rates in patients.

\textbf{Contrast Learning.} Contrastive learning \cite{1640964} is an unsupervised pre-training method that learns similar/dissimilar representations from data that are organized into similar/dissimilar pairs. These methods \cite{oord2018representation,henaff2020data,wu2018unsupervised,he2020momentum,misra2020self} mainly calculate contrast objective function at feature-level to learn good representations under unsupervised training. In this paper, we adapt it to supervised training through object-level contrast.

\section{METHODOLOGY}
\begin{figure*}[t]
    \centering
    \includegraphics[width=0.82\textwidth]{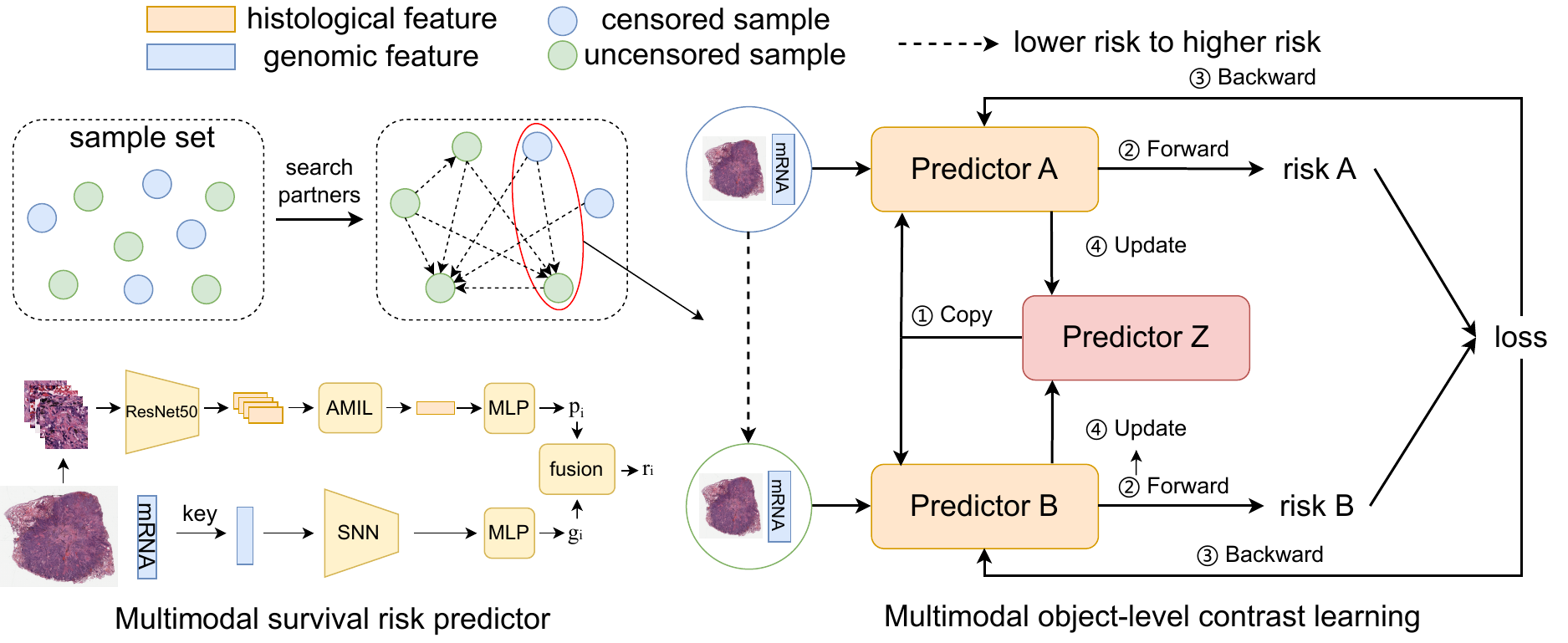}
    \caption{\textbf{Architecture of mutimodal object-level contrast learning and multimodal survival risk predictor.}}
    \label{fig2}
\end{figure*}

In this paper, we construct a multimodal survival risk predictor and propose a multimodal object-level constrast learning method, the overall architecture of our proposed method is illustrated in Figure \ref{fig2}. In the training phase, we first construct contrast pairs based on the 
grount-truth relative risk relationship among patients in the training set. 
Then, the survival risk predictions of the each modality is obtained through the multi-modal survival risk predictor, and the survival risk predictor is updated through the cross-modal object-level contrast learning objective. 
In the inference stage, the patient's multimodal data is passed through the survival risk predictor to obtain the patient's survival risk predictions for each modality, and finally the patient's final prediction is obtained through decision-level fusion. 

\subsection{Problem Formulation}
Consider a set of $N$ samples, $s_i$ represent the $i$-th sample. We have a tuple $\{Pa_i, Ge_i, d_i, o_i, c_i\}$ for sample $s_i$, where $Pa_i$ represents sample's pathological image, $Ge_i$ represents sample's genomics data, $c_i$ represent the censoring status (1 for uncensored and 0 for censored), $d_i$ represent the time elapsed between diagnosis and death and $o_i$ represent the time elapsed between diagnosis and last follow up. Then we can define the $t_i$ that is either equal to $d_i$ ($c_i=0$) or $o_i$ ($c_i=1$).
For sample $s_i$, we now have a tuple $\{Pa_i, Ge_i, t_i, c_i\}$ , and we want to obtain a survival risk predictor $h(x,y)$ to predict a survival risk $r_i = h(Pa_i,Ge_i)$. A good survival risk predictor should precisely predict the concordance of survival risk between samples, which means for any two samples $s_i$ and $s_j$, if $t_i$ is bigger than $t_j$, the $r_i$ should smaller than $r_j$.

% \onecolumn
% \twocolumn

\subsection{Construct Contrast Learning Pairs}
In the training sample sets, each sample will search for its learning partners, and each pair of learning partners constitutes a contrast learning pair. 
For each uncensored sample $s_i$ in the training sample set, $s_i$'s partners can be every uncensored patients who have survived longer or shorter than $s_i$. For each censored patient $s_j$ in the patient set, $s_j$'s learning partners can only be uncensored patients who have survived for a shorter period of time than $s_j$'s censored time, because the relative risks between such pairs of patients are clear.
In a contrast learning pair,  for the patients $s_i$ and $s_j$, if $t_i<t_j$, $s_i$ is regarded as a reference with higher risk (end point of dotted arrow in Figure \ref{fig2}), and $s_j$ is regarded as a reference with lower risk (start point of dotted arrow in Figure \ref{fig2}).

\subsection{Multimodal Survival Risk Predictor}
We construct a survival risk predictor for each modality respectively. In the training phase, the cross-modal object-level contrast learning is used for training survival risk predictor, and in the inference phase, decision-level fusion is used to obtain the final prediction. The survival risk predictor for each modality and decision-level fusion are described below.

\textbf{Pathological images.} Pathological images are often gigapixel images and lack fine-grained annotations, making it difficult to directly apply deep learning methods. It is usually considered as a weakly supervised learning problem and multiple instance learning methods are used to assist diagnosis. First, we tile all pathological image $Pa_i$ into nonoverlapping small patches $\{sp_{i,1},sp_{i,2},...,sp_{i,n_i}\}$, $n_i$ is the number of patches cut out of $Pa_i$. Then, we use a pretrained ResNet50 \cite{resnet} to extract all patches' features $pf_{i,j}$. We use attention-based multiple instance learning (AMIL) \cite{amil, yao2020whole, 10.1007/978-3-030-32239-7_55, DSMIL} to assign a dynamic weight $\alpha_{i,j}$ to each patch-level feature and aggregate them into pathological-level features $wf_i$:
\begin{equation}
    wf_i = \sum_{j=1}^{n_i}\alpha_{i,j}pf_{i,j}
\end{equation}
\begin{equation}
    \alpha_{i,j} = \frac{exp(w^T(tanh(V pf_{i,j}^T)\odot sigma(U pf_{i,j}^T)))}{\sum_{k=1}^{n_i}exp(w^T(tanh(V pf_{i,k}^T)\odot sigma(U pf_{i,k}^T)))}
\end{equation}
The weight of each patch feature $\alpha_{i,j}$ is parameterized by $U$ and $V$, which can be learned through backpropagation.
We use pathological-level features as the input of MLP for survival risk prediction, and use the output of the last layer of MLP with a sigmoid activation function to obtain survival risk prediction $p_i$.

\textbf{Genomics.} Genomics have hundreds to thousands of features with relatively few training samples, traditional artificial neural networks are prone to overfitting.
Self-Normalizing Network (SNN) \cite{klambauer2017self} employ more robust regularization techniques on high-dimensional low sample size genomics data and has been demonstrated to be effective by \cite{chen2020pathomic,chen2021multimodal,chen2022pan, klambauer2017self}. We first filter out molecular features related to cancer biology, including transcription factors, tumor suppression, cytokines and growth factors, cell differentiation markers, homeodomain proteins, translocated cancer genes, and protein kinases \cite{subramanian2005gene, liberzon2015molecular} from all mRNA data. 
We then use SNN with sigmoid activation to obtain the final prediction $g_i$.

\textbf{Decision-level fusion.}
The network is trained using object-level contrast learning, and it's natural to choose decision-level fusion in the inference phase.
In order to reduce the variance of the prediction, in the inference phase, we use the mean of the two modalities' prediction as the final survival risk prediction $r_i = \frac{p_i+g_i}{2}$.

\subsection{Multimodal Object-level Contrast Learning}
The core idea of our method is to use contrast learning pairs to train a precise survival risk predictor. The contrast learning method we proposed takes the previously constructed contrast learning pairs as the input. The model maintains three risk predictors $P_A$, $P_B$, and $P_Z$ simultaneously. The initial parameters of $P_A$ and $P_B$ are copied from the parameters of $P_Z$. In the training phase, samples in contrast learning pair get their risk predictions from $P_A$ and $P_B$ respectively and update $P_A$ and $P_B$ by referring to each other's prediction.

\textbf{Object-level contrast learning.} In a forward iteration, the lower risk sample and the higher risk sample get their predicted survival risk value $r_A$ and $r_B$ by $P_A$ and $P_B$, respectively. The object-level contrast learning objective we propose has the following form:
\begin{equation}\label{eq5}
    \min\, loss = \frac{r_A}{r_B}
\end{equation}
Then the gradient of the output neurons of $P_A$ and $P_B$ has:
\begin{equation}
    \frac{\partial loss}{\partial O_A} = \frac{\partial loss}{\partial r_A}\cdot \frac{\partial r_A}{\partial O_A}=\frac{r_A}{r_B}(1-r_A)  
\end{equation}
\begin{equation}
    \frac{\partial loss}{\partial O_B} = \frac{\partial loss}{\partial r_B}\cdot \frac{\partial r_B}{\partial O_B}=-\frac{r_A}{r_B}(1-r_B)  
\end{equation}
The term on the left indicates that the two predictors have gradients of the same scale in different directions, similar to the Cox partial likelihood method. The right term is an term specific to our method. When $r_A$ is bigger than $r_B$, the risk relationship for this sample pair is mispredicted, with $\frac{r_A}{r_B}$ greater than one having a larger gradient, and $(1-r_A)$ less than $(1-r_B)$ resulting in a larger gradient for uncensored higher risk sample. On the contrary, when the risk relationship of the sample pair is correctly predicted, lower risk sample who's ground truth risk value may be smaller than $r_A$ will have a larger gradient.

\textbf{Multimodal object-level contrast learning.} We further extend our method to multimodal scenarios. In a forward step, each sample in the contrast learning sample pair use the survival risk predictor to obtain prediction $p_A$ and $p_B$ based on pathological images and prediction $g_A$ and $g_B$ based on genes. Then we have the cross-modal object-level contrast learning objective:
\begin{equation}
    \min\, loss = \frac{p_A}{p_B}+\frac{p_A}{g_B}+\frac{g_A}{p_B}+\frac{g_A}{g_B}+\frac{p_A+g_A}{p_B+g_B}
\end{equation}
This simultaneously utilizes the mutual inspiration from intra-modal and inter-modal of contrast learning pairs to train the multimodal survival risk predictor. The pseudo-code for MOL is shown in Algorithm \ref{alg1}.
\begin{algorithm}
	%\textsl{}\setstretch{1.8}
	\renewcommand{\algorithmicrequire}{\textbf{Input:}}
	\renewcommand{\algorithmicensure}{\textbf{Output:}}
	\caption{Multimodal object-level contrast learning}
	\label{alg1}
	\begin{algorithmic}[1]
            \REQUIRE Training sample set $\{Pa_i, Ge_i, t_i, c_i\}$, $i=1,2,...,N$
            \ENSURE  Survival risk predictor's parameter $\theta_Z$
            \STATE Construct contrast learning pairs in the training sample set. $\{\{Pa_{i,A},Ge_{i,A},t_{i,A},c_{i,A}\},\{Pa_{i,B},Ge_{i,B},t_{i,B},c_{i,B}\}\}$, $i=1,2,...,M$
		\STATE Initialization: $i\leftarrow 0$, $\theta_Z\leftarrow\theta_0$
		\REPEAT
		      \STATE $i \leftarrow i + 1$
                \STATE Copy $\theta_Z$ to $\theta_A$ and $\theta_B$
                \STATE Forward propagation to get predicted risk: $p_A,g_A\leftarrow P_A(Pa_{i,A},Ge_{i,A})$ and $p_B,g_B\leftarrow P_B(Pa_{i,B},Ge_{i,B})$
                \STATE Computate the loss function: $loss\leftarrow \frac{p_A}{p_B}+\frac{p_A}{g_B}+\frac{g_A}{p_B}+\frac{g_A}{g_B}+\frac{p_A+g_A}{p_B+g_B}$.
                \STATE Backward propagation to update $\theta_A$ and $\theta_B$.
                \STATE Update the $\theta_Z\leftarrow\frac{\theta_A+\theta_B}{2}$
		\UNTIL $i==M$  		
	\end{algorithmic}  
\end{algorithm}

\section{Experiments and Results}
\label{sec:experiment}

\subsection{Datasets \& Evaluation Metrics}
% \textbf{Datasets \& Evaluation Metrics.}
The Cancer Genome Atlas (TCGA) is a public cancer data consortium that contains matched diagnostic pathological images and genomic data with labeled survival times and censorship statuses. We use the commonly used Lung Adenocarcinoma (LUAD) and Kidney Renal Clear Cell Carcinoma (KIRC) to validate the effectiveness our method. For each patient, we select the primary tumor samples from the same center, filter out the samples without complete survival time or censor time, and finally get 957 patients with both pathological images and RNA-Seq data, including 451 patients with LUAD and 506 patients with KIRC. 
% The detail patients number of each modality is shown in Table \ref{tab0}. 
We totally get 972 WSIs. All pathological images are cut into non-overlapping 256x256 patches at a 20x magnification, and  patches with large non-tissue areas are deleted. Finally, each pathological image have an average of 13173 patches. We train our proposed method in a 5-fold cross-validation, and used the concordance index (C-index) \cite{wang2019machine} to measure the predictive performance on two datasets, respectively.

\subsection{Implementation Details}
% \textbf{Implementation Details.}
We have two hidden layers of genomic data, each with a neuron count of 256. For pathological data, there are also two hidden layers, and the number of neurons is 512 and 256 respectively. The Adam \cite{kingma2017adam} optimizer is used to optimize the network parameters with a fixed learning rate of 0.0002. Due to the number of patches of each WSI is always inconsistent, we set batch size to 1, and adopt gradient accumulation to achieve the same effect, backward propagation is carried out every 128 times of forward propagation. We employ dropout \cite{srivastava2014dropout} technology to mitigate overfitting, and set the dropout rate at 0.25. Random seed value is fixed to 1. All experiments are done on Nvidia 1080 GPUs. A 5-fold cross-validation was used on all datasets for all models, and the mean C-index of the validation set in five-fold are used for comparison.
% A total of 100 models were trained on two datasets to demonstrate the effectiveness of our proposed method.

\begin{table}[tbp]
    \centering
    \caption{The C-index (mean$\pm$std) performance of MOC against several state-of-the-art deep learning-based methods.}
    \label{tab1}
    \begin{threeparttable}
        \begin{tabular}{cccccc}
            \hline
            \multirow{2}{*}{\textbf{Model}} & \multicolumn{2}{c}{\textbf{Modality}} & \multirow{2}{*}{\textbf{LUAD}} & \multirow{2}{*}{\textbf{KIRC}} \\ \cline{2-3}
                                            & P                 & G                 &                                &                                \\ \hline
            % SNN\cite{chen2020pathomic}      &                   & \checkmark        & 0.591$\pm$0.044                    & 0.697$\pm$0.029            \\ \hline
            % Deep Sets\cite{zaheer2017deep}  & \checkmark        &                   & 0.498$\pm$0.004                    & 0.500$\pm$0.000            \\
            % AMIL\cite{amil}                 & \checkmark        &                   & 0.561$\pm$0.022                    & 0.578$\pm$0.030            \\ \hline
            CoxPH \cite{cox1972regression}  &                   & \checkmark        & 0.531$\pm$0.082                & 0.550$\pm$0.070                \\ \hline
            DeepSurv \cite{katzman2018deepsurv} & \checkmark    &                   & 0.534$\pm$0.077                & 0.620$\pm$0.043                \\ \hline
            MCAT\cite{chen2021multimodal}   & \checkmark        & \checkmark        & 0.612$\pm$0.061                    & 0.666$\pm$0.027            \\
            PORPOISE\cite{chen2022pan}      & \checkmark        & \checkmark        & 0.568$\pm$0.079                    & 0.668$\pm$0.023            \\ 
            PONET-OH\cite{qiu2023deep}      & \checkmark        & \checkmark        & 0.618$\pm$0.049                    & 0.695$\pm$0.043            \\ 
            \textbf{MOC(ours)}            & \checkmark        & \checkmark        & \textbf{0.655$\pm$0.029}          & \textbf{0.701$\pm$0.012}   \\ \hline
        \end{tabular}
        \begin{tablenotes}
            \footnotesize
            \item P represent pathological images, G represent genomic data.
        \end{tablenotes}
    \end{threeparttable}
\end{table}

\begin{figure*}[tbp]
    \centering
    \includegraphics[width=0.8\textwidth]{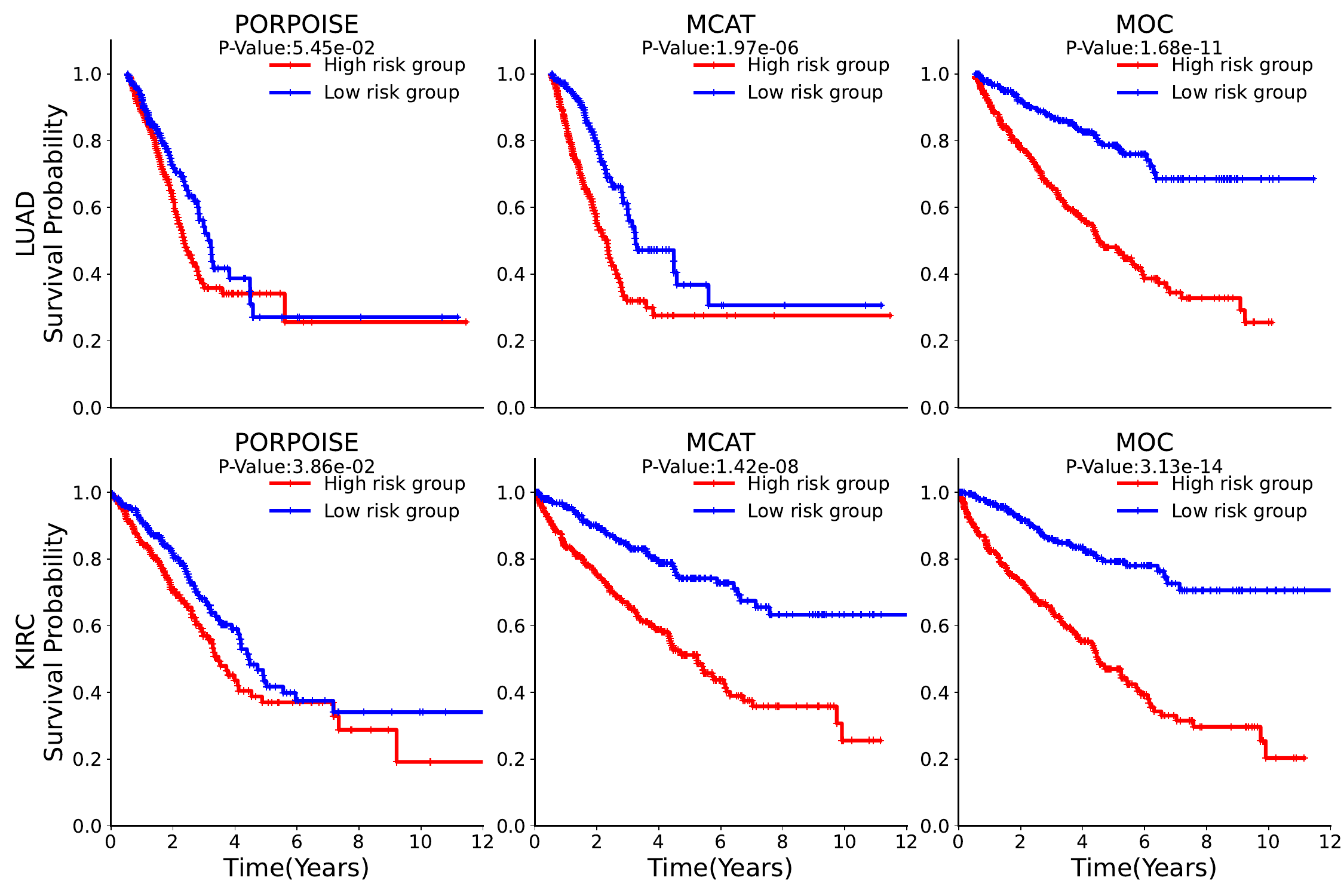}
    \caption{\textbf{Kaplan-Meier Curve of different Models}.
    In each cancer, patients are divided into high risk group and low risk group based on the median survival risk predictions of different models. Log-rank test is used to test for statistical significance in survival distributions between low- and high-risk patients
    }
    \label{fig3}
\end{figure*}

\subsection{Results}
\textbf{Compare with State-of-the-art.} 
For comparison of other works, we implement and compare with several state-of-the-art methods used for survival risk prediction. Including unimodal methods \cite{cox1972regression, katzman2018deepsurv} and multimodal methods \cite{chen2021multimodal, chen2022pan, qiu2023deep}.
% SNN \cite{klambauer2017self} is a unimodal method for genomic features, which has been used previously for survival risk prediction by Chen et al. \cite{chen2020pathomic}. Deep Sets \cite{zaheer2017deep} proposed sum pooling over instance-level features to get bag-level features, which is one of the first deep learning methods for multiple instance learning. AMIL \cite{amil} use attention based pooling to aggregate instance-level features, and it has been explored by many works \cite{yao2020whole, 10.1007/978-3-030-32239-7_55, DSMIL}.
% MMF \cite{chen2022pan} is a multimodal method using genomic and pathological features for cancer survival risk prediction, which fuses multimodal features by concatenating in the feature-level and then predict the survival risk using joint features. MCAT \cite{chen2021multimodal} introduced the co-attention mechanism \cite{yu2019deep} to fuse genomic and histologial features. Their inspiration is to learn how histology patches attend to genes when predicting patient survival, and their method achieved state-of-the-art performance on multiple cancer datasets.
For multimodal dataset, we compare MSRP trained by MOC with PORPOISE, MCAT and PONET, which are the state-of-the-art for multimodal cancer surivival risk prediction.
The results are shown in Table \ref{tab1}. It can be seen that our method achieves best performance than all other methods. C-index has been improved by at least 6\% on LUAD and 0.8\% on KIRC than the state-of-the-art methods. 
The comparison of Kaplan-Meier Curve of different models is shown in Figure \ref{fig3}, patients are divided into high risk group and low risk group based on the median survival risk predictions. Log-rank test is used to test for statistical significance in survival distributions between low risk group and high risk group. In LUAD, the p-value of PORPOISE and MCAT are greater than 0.05, indicating PORPOISE and MCAT failed to distinguish different survival risk level. On the contrary, our method's p-value is small than 0.05, indicating our method succeed to distinguish different survival risk level. In KIRC, all of PORPOISE, MCAT and MOC succeed to distinguish different survival risk level, and our method has smaller p-value than the others.

\begin{table}[tbp]
    \centering
    \caption{Evaluation of different training method for survival risk predicrion on pathological images by C-index (mean$\pm$std).}
    \label{tab2}
    \begin{tabular}{ccc}
         \hline
        Training Method        & LUAD                 & KIRC                 \\ \hline
        AMIL-ELB \cite{steck2007ranking}           & 0.591$\pm$0.044          & 0.626$\pm$0.043        \\
        AMIL-NLL \cite{zadeh2020bias}            & 0.561$\pm$0.022          & 0.578$\pm$0.030          \\
        \textbf{AMIL-OC} & \textbf{0.634$\pm$0.057} & \textbf{0.634$\pm$0.039} \\ \hline
    \end{tabular}
\end{table}

\textbf{Ablation study of training method for survival risk predicrion.}
We validate the effectiveness of our proposed OC on pathological images and genomic data respectively. On the pathological images, we extend the semi-parametric method based on Cox partial likelihood previously proposed by Steck \cite{steck2007ranking} to deep learning (ELB). Then, we train the same model AMIL, using different training methods ELB, NLL and OC to compare their performances. On the genomics data, we train the same model SNN, using different training methods CoxPH, NLL and OC to compare their performance. The experimental results on two cancer datasets are shown in Table \ref{tab2} and Table \ref{tab3}. It can be seen that the OC achieves performance that is either comparable or surpassing the best on both datasets, including pathological images and genomics data.

\begin{table}[thbp]
    \centering
    \caption{Evaluation of different training method for survival risk predicrion on genomics data by C-index (mean$\pm$std).}
    \label{tab3}
    \begin{tabular}{ccc}
         \hline
        Training Method        & LUAD                 & KIRC                 \\ \hline
        SNN-CoxPH \cite{cox1972regression}           & 0.507$\pm$0.082          & 0.518$\pm$0.094        \\
        SNN-NLL \cite{zadeh2020bias}            & 0.591$\pm$0.044          & \textbf{0.697$\pm$0.029}          \\
        \textbf{SNN-OC} & \textbf{0.637$\pm$0.037} & 0.688$\pm$0.012 \\ \hline
    \end{tabular}
\end{table}

\textbf{Ablation study of modality.}
We compare the multimodal model trained by MOC with the unimodal models trained by OC on LUAD and KIRC datasets. The results are shown in Table \ref{tab4}. It can be seen that the performance of MOC outperform the OC in any of the modalities and multimodal without cross-modal cooperation. This demonstrates that cross-modal contrast learning can capture meaningful inter-modal information to enhance survival risk prediction.

\begin{table}[thbp]
    \centering
    \caption{Ablation study of modality. C-index (mean$\pm$std) is used for evaluation.}
    \label{tab4}
    \begin{tabular}{ccc}
         \hline
        Modality                & LUAD                 & KIRC        \\ \hline
        Genomic                 & 0.637$\pm$0.037          & 0.688$\pm$0.012 \\
        Pathological            & 0.634$\pm$0.057          & 0.634$\pm$0.039 \\
        Multimodal(w/o cross-modal) & 0.601$\pm$0.035 & 0.696$\pm$0.032 \\
        Multimodal(w/ cross-modal) & \textbf{0.655$\pm$0.029} & \textbf{0.701$\pm$0.012} \\ \hline
    \end{tabular}
\end{table}

\section{Conclusions}
In this paper, we propose the object-level contrast learning (OC) for cancer surivival risk prediction and we demonstrate its effectiveness in pathological images and genomics data on two cancers datasets. We further extend OC to multimodal scenarios, proposing multimodal object-level contrast learning (MOC). MOC adopts employs cross-modal contrast to train the multimodal survival risk predictor, and we demonstrate that cross-modal contrast can capture inter-modal information useful for survival risk prediction. Finally, the multimodal survival risk predictor trained by MOC achieves the state-of-the-art performance on two multimodal cancer survival risk prediction datasets.
\begin{acks}
This work is supported by the National Natural Science Foundation of China (62276250), the National Key R\&D Program of China (2022YFF1203303).
\end{acks}

\bibliographystyle{ACM-Reference-Format}
\balance
\bibliography{refs}

\end{document}